# Object Detection in Autonomous Vehicles: Status and Open Challenges

Abhishek Balasubramaniam and Sudeep Pasricha
Colorado State University

*Abstract*— Object detection is a computer vision task that has become an integral part of many consumer applications today such as surveillance and security systems, mobile text recognition, and diagnosing diseases from MRI/CT scans. Object detection is also one of the critical components to support autonomous driving. Autonomous vehicles rely on the perception of their surroundings to ensure safe and robust driving performance. This perception system uses object detection algorithms to accurately determine objects such as pedestrians, vehicles, traffic signs, and barriers in the vehicle's vicinity. Deep learning-based object detectors play a vital role in finding and localizing these objects in real-time. This article discusses the state-of-the-art in object detectors and open challenges for their integration into autonomous vehicles.

## I. INTRODUCTION

Autonomous vehicles (AVs) have received immense attention in recent years, in large part due to their potential to improve driving comfort and reduce injuries from vehicle crashes. It has been reported that more than 36,000 people died in 2019 due to fatal accidents on U.S. roadways [1]. AVs can eliminate human error and distracted driving that is responsible for 94% of these accidents [2]. By using sensors such as cameras, lidars, and radars to perceive their surroundings, AVs can detect objects in their vicinity and make real-time decisions to avoid collisions and ensure safe driving behavior.

AVs are generally categorized into six levels by the SAE J3016 standard [3] based on their extent of supported automation (see Table 1). While level 0 – 2 vehicles provide increasingly sophisticated support for steering and acceleration, they heavily rely on the human driver to make decisions. Level 3 vehicles are equipped with Advanced Driver Assistance Systems (ADAS) to operate the vehicle in various conditions, but human intervention may be requested to safely steer, brake, or accelerate as needed. Level 4 vehicles are capable of full self-driving mode in specific conditions but will not operate if these conditions are not met. Level 5 vehicles can drive without human interaction under all conditions.

Automotive manufactures have been experimenting with AVs since the 1920s. The first modern AV was designed as part of CMU NavLab's autonomous land vehicle project in 1984 with level 1 autonomy that was able to steer the vehicle while the acceleration was controlled by a human driver [4]. This was followed by an AV designed by Mercedes-Benz in 1987 with level 2 autonomy that was able to control steering and acceleration with limited human supervision [5]. Subsequently, most major auto manufacturers such as General Motors, Bosch, Nissan, and Audi started to work on AVs.

Table 1: SAE J3016 levels of automation

| SAE Level | Name | Driving Environment Monitor |
|---|---|---|
| 0 | No Automation | Human Driver |
| 1 | Driver Assistance | |
| 2 | Partial Driving Automation | |
| 3 | Conditional Driving Automation | ADAS System |
| 4 | High Driving Automation | |
| 5 | Full Driving Automation | |

Tesla was the first company to commercialize AVs with their Autopilot system in 2014 that offered level 2 autonomy [6]. Tesla AVs were able to travel from New York to San Francisco in 2015 by covering 99% of the distance autonomously. In 2017, Volvo launched their Drive Me feature with level 2 autonomy, with their vehicles traveling autonomously around the city of Gothenburg in Sweden under specific weather conditions [7]. Waymo has been testing its AVs since 2009 and has completed 200 million miles of AV testing. They also launched their driverless taxi service with level 4 autonomy in 2018 in the metro Phoenix area in USA with 1000 – 2000 riders per week, among which 5 – 10% of the rides were fully autonomous without any drivers [8]. Cruise Automation started testing a fleet of 30 vehicles in San Francisco with level 4 autonomy in 2017, launched their self-driving Robotaxi service in 2021 [9]. Even though Waymo and Cruise support level 5 autonomy, their AVs are classified as level 4 because there is still no guarantee that they can operate safely in all weather and environmental conditions.

AVs rely heavily on sensors such as cameras, lidars, and radars for autonomous navigation and decision making. For example, Tesla AVs rely on camera data with six forward facing cameras and ultrasonic sensors. In contrast, Cruise AVs use a sensor cluster that consists of a radar in the front while camera and lidar sensors are mounted on the top of the AV to provide a 360-degree view of the vehicle surroundings [9]. One of the main tasks involved in achieving robust environmental perception in AVs is to detect objects in the AV vicinity using software-based object detection algorithms. Object detection is a computer vision task that is critical for recognizing and localizing objects such as pedestrians, traffic lights/signs, other vehicles, and barriers in the AV vicinity. It is the foundation for high-level tasks during AV operation, such as object tracking, event detection, motion control, and path planning.

The modern evolution of object detectors began 20 years ago with the Viola Jones detector [10] used for human face detection in real-time. A few years later, Histogram of Oriented Gradient (HOG) [11] detectors became popular for pedestrian detection. HOG detectors were then extended to Deformable Part-based Models (DPMs), which were the first models to focus on multiple object detection [12]. With growing interest in deep neural networks around 2014, the Regions with Convolutional Neural Network (R-CNN) deep neural network model led to a breakthrough for multiple object detection, with a 95.84% improvement in Mean Average Precision (mAP) over the state-of-the-art. This development helped redefine the efficiency of object detectors and made them attractive for entirely new application domains, such as for AVs. Since 2014, the evolution in deep neural networks and advances in GPU technology have paved the way for faster and more efficient object detection on real-time images and videos [10]. AVs today rely heavily on these improved object detectors for perception, pathfinding, and other decision making.

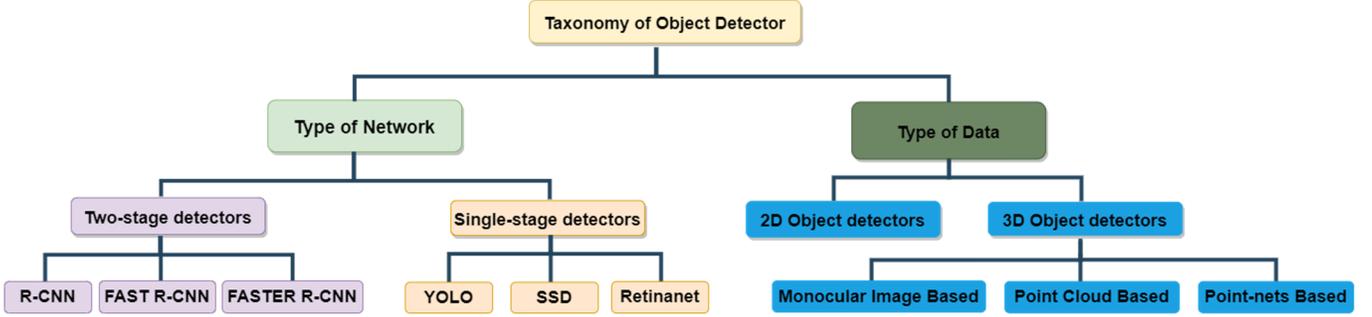

**Figure 1:** Taxonomy of object detectors

This article discusses contemporary deep learning based object detectors, their usage, optimization, and limitations for AVs. We also discuss open challenges and future directions.

## II. Overview of Object Detectors

Object detection consists of two sub-tasks: localization, which involves determining the location of an object in an image (or video frame), and classification, which involves assigning a class (e.g., 'pedestrian', 'vehicle', 'traffic light') to that object. Figure 1 illustrates a taxonomy of state-of-the-art deep learning-based object detectors. We discuss the taxonomy of these object detectors in this section.

### A. Two-stage vs Single stage object detectors

Two-stage deep learning based object detectors involve a two-stage process consisting of 1) region proposals and 2) object classification. In the region proposal stage, the object detector proposes several Regions of Interest (ROIs) in an input image that have a high likelihood of containing objects of interest. In the second stage, the most promising ROIs are selected (with other ROIs being discarded) and objects within them are classified [13]. Popular two-stage detectors include R-CNN, Fast R-CNN, and Faster R-CNN. In contrast, single-stage object detectors use a single feed-forward neural network that creates bounding boxes and classifies objects in the same stage. These detectors are faster than two-stage detectors but are also typically less accurate. Popular single-stage detectors include YOLO, SSD, EfficientNet, and RetinaNet.

Figure 2 illustrates the difference between the two types of object detectors. Both types of object detectors are typically evaluated using the mAP and Intersection over Union (IoU) accuracy metrics. mAP is the mean of the ratio of precision to recall for individual object classes, with a higher value indicating a more accurate object detector. IoU measures the overlap between the predicted bounding box and the ground truth bounding box. Formally, IoU is the ratio of the area of overlap between the (bounding and ground truth) boxes and the area of union between the boxes. Figure 3 illustrates the IoU of an object detector prediction and the ground truth. Figure 3(a) shows a highly accurate IoU and 3(b) shows a less accurate IoU.

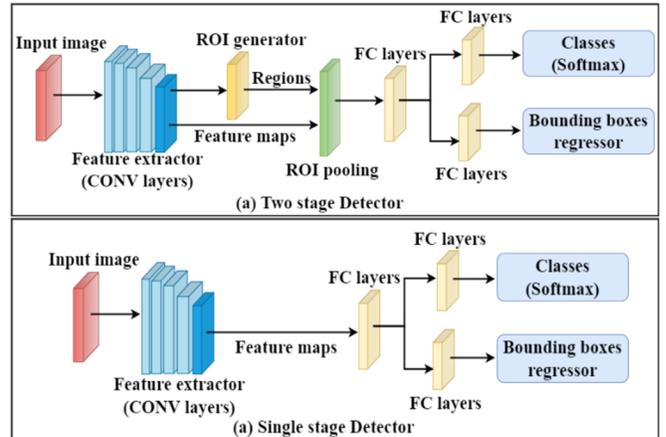

**Figure 2**: Two-stage vs Single stage object detector diagram

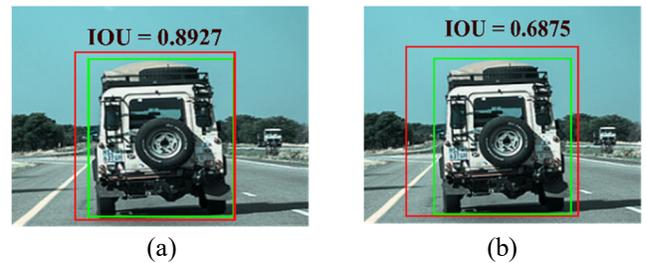

**Figure 3**: Example of an IOU; green box: ground truth; red box: prediction

R-CNN was one of the first deep learning-based object detectors and used an efficient selective search algorithm for ROI proposals as part of a two-stage detection [13]. Fast R-CNN solved some of the problems in the R-CNN model, such as low inference speed and accuracy. In the Fast R-CNN model, the input image is fed to a Convolutional Neural Network (CNN), generating a feature map and ROI projection. These ROIs are then mapped to the feature map for prediction using ROI pooling. Unlike R-CNN, instead of feeding the ROI as input to the CNN layers, Fast R-CNN uses the entire image directly to process the feature maps to detect objects [14]. Faster R-CNN used a similar approach to Fast R-CNN, but instead of using a selective search algorithm for the ROI proposal, it employed a separate network that fed the ROI to the ROI pooling layer and the feature map, which were then reshaped and used for prediction [15].

Single-stage object detectors such as YOLO (You only look once) are faster than two-stage detectors as they can predict objects on an input with a single pass. The first YOLO variant, YOLOv1, learned generalizable representations of objects to detect them faster [16]. In 2016, YOLOv2 improved upon YOLOv1 by adding batch normalization, a high-resolution classifier, and use of anchor boxes to create bounding boxes instead of using a fully connected layer like YOLOv1 [17]. In 2018, YOLOv3 was proposed with a 53 layered backbone-based network that used an independent logistic classifier and binary cross-entropy loss to predict overlapping bounding boxes and smaller objects [18]. Single-Shot Detector (SSD) models were proposed as a better option to run inference on videos and real-time applications as they share features between the classification and localization task on the whole image, unlike YOLO models that generate feature maps by creating grids within an image. While the YOLO models are faster than SSD, they trail behind SSD models in accuracy [19]. Even though YOLO and SSD models provide good inference speed, they have a class imbalance problem when detecting small objects. This issue was addressed in the RetinaNet detector that used a focal loss function during training and a separate network for classification and bounding box regression [20].

In 2020, YOLOv4 introduced two important techniques: 'bag of freebies' which involves improved methods for data augmentation and regularization during training and 'bag of specials' which is a post processing module that allows for better mAP and faster inference [21]. YOLOv5, which was also introduced in 2020, proposed further data augmentation and loss calculation improvements. It also used auto-learning bounding box anchors to adapt to a given dataset [22]. Another variant called YOLOR (You Only Learn One Representation) was proposed in 2021 and used a unified network that encoded implicit and explicit knowledge to predict the output. YOLOR can perform multitask learning such as object detection, muti-label image classification, and feature embedding using a single model [23]. The YOLOX model, also proposed in 2021, uses an anchor-free, decoupled head technique that allows the network to process classification and regression using separate networks. Unlike the YOLOv4 and YOLOv5 models, YOLOX has reduced number of parameters and increased inference speed [24]. The performance of each model in terms of mAP and inference speed is summarized in Table 2.

*B. 2D vs 3D object detectors*

2D object detectors typically use 2D image data for detection, but recent work has also proposed a sensor-fusion based 2D object detection approach that combines data from a camera and radar [25]. 2D object detectors provide bounding boxes with four Degrees of Freedom (DOF). Figure 4 shows the most common approach for encoding bounding boxes 4(a): [x, y, height, width] and 4(b): [xmin, ymin, xmax, ymax] [26]. Unfortunately, 2D object detection can only provide the position of the object on a 2D plane but does not provide information about the depth of the object. Depth of the object is important to predict the shape, size, and position of the object to enable improved performance in various self-driving tasks such as path planning, collision avoidance, etc.

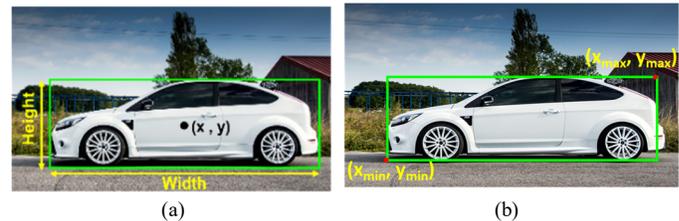

(a)      (b)

**Figure 4**: Commonly used bounding box encoding methods

**Table 2: 2D and 3D object detector models and their performance**

| Name | Year | Type | Dataset | mAP | Inference rate (fps) |
|---|---|---|---|---|---|
| R-CNN [13] | 2014 | 2D | Pascal VOC | 66% | 0.02 |
| Fast R-CNN [14] | 2015 | 2D | Pascal VOC | 68.8% | 0.5 |
| Faster R-CNN [15] | 2016 | 2D | COCO | 78.9% | 7 |
| YOLOv1 [16] | 2016 | 2D | Pascal VOC | 63.4% | 45 |
| YOLOv2 [17] | 2016 | 2D | Pascal VOC | 78.6% | 67 |
| SSD [19] | 2016 | 2D | Pascal VOC | 74.3% | 59 |
| RetinaNet [20] | 2018 | 2D | COCO | 61.1% | 90 |
| YOLOv3 [18] | 2018 | 2D | COCO | 44.3% | 95.2 |
| YOLOv4 [21] | 2020 | 2D | COCO | 65.7% | 62 |
| YOLOv5 [22] | 2021 | 2D | COCO | 56.4% | 140 |
| YOLOR [23] | 2021 | 2D | COCO | 74.3% | 30 |
| YOLOX [24] | 2021 | 2D | COCO | 51.2% | 57.8 |
| Complex-YOLO [27] | 2018 | 3D | KITTI | 64.00% | 50.4 |
| Complexer-YOLO [28] | 2019 | 3D | KITTI | 49.44% | 100 |
| Wen et al. [29] | 2021 | 3D | KITTI | 73.76% | 17.8 |
| RAANet [30] | 2021 | 3D | NuScenes | 62.0% | - |

Figure 5 shows the difference between a 2D and 3D object detector output on real images. 3D object detectors use data from a camera, lidar, or radar to detect objects and generator 3D bounding boxes. These detectors provide bounding boxes with (x, y, z) and (height, width, length) along with yaw information [26]. These object detectors use several approaches, such as point clouds and frustum pointnets, for predicting objects in real-time. Point cloud networks can directly use 3D data, but the complexity and cost of computing are very high, so some networks use 2D to 3D lifting while compensating for the loss of information. Pointnets are used along with RGB images, where 2D bounding boxes are obtained using RGB images. Then these boxes are used as ROIs for 3D object detection which reduces the search effort [26]. Monocular image-based methods have also been proposed that use an RGB image to predict objects on the 2D plane and then perform 2D to 3D lifting to create 3D object detection results.

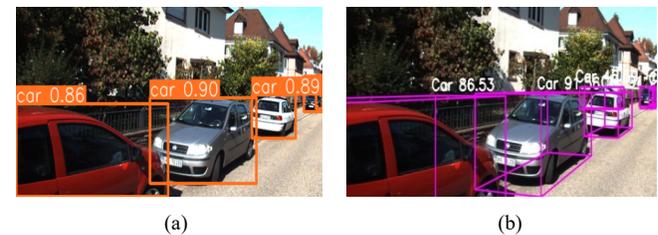

(a)      (b)

**Figure 5**: Object detection modalities: (a) 2D vs. (b) 3D

Recent years have seen growing interest in 3D object detection with deep learning. Complex-YOLO, an extension of YOLOv2, used a Euler Region Proposal Network (E-RPN), based on an RGB Birds-Eye-View (BEV) map from point cloud data to get 3D proposals. The network exploits the YOLOv2 network followed by E-RPN to get the 3D proposal [27]. Later in 2019, Complexer-YOLO achieved semantic segmentation and 3D object detection using Random Finite Set (RFS) [28]. The more recent work on 3D object detection by Wen et al. [29] in 2021 proposed a lightweight 3D object detection model that consists of three submodules: 1) point transform module, which extracts point features from the RGB image based on the raw point cloud, 2) voxelization, which divides the features into equally spaced voxel grids and then generates a many-to-one mapping between the voxel grids and the 3D point clouds, and 3) point-wise fusion module, which fuses the features using two fully connected layers. The output of the point-wise fusion module is encoded and used as input for the model. Another 3D detector proposed in 2021 called RAANet used only lidar data to achieve 3D object detection [30]. It used the BEV lidar data as input for a region proposal network which was then used to create shared features. These shared features were used as the input for an anchor free network to detect 3D objects. The performance of these models is summarized in Table 2.

### III. Deploying Object Detectors in AVs

Deploying deep learning-based object detector models in AVs has its own challenges, mainly due to the resource-constrained nature of the onboard embedded computers used in vehicles. These computing platforms have limited memory availability and reduced processing capabilities due to stringent power caps, and high susceptibility to faults due to thermal hotspots and gradients, especially during operation in the extreme conditions found in vehicles. As the complexity of the object detector model increases, the memory and computational requirements, and energy overheads also increase. In this section, we discuss techniques to improve object detector model deployment efficiency. The performance of some of the latest works on this topic is summarized in Table 3.

#### A. Pruning

Pruning a neural network model is a widely used method for reducing the model's memory footprint and computational complexity. Pruning was first used in the 1990s to reduce neural network sizes for deploying them on embedded platforms [31]. Pruning involves removing redundant weights and creating sparsity in the model by training the model with various regularization techniques (L1, L2, unstructured, and structured regularization). Sparse models are easier to compress, and the zero weights created during pruning can be skipped during inference, reducing inference time, and increasing efficiency. While most pruning approaches target deep learning models for the simpler image classification problem, relatively fewer works have attempted to prune the more complex object detector models. Wang et al. [32] proposed using a channel pruning strategy on SSD models in which they start by creating a sparse normalization and then prune the channels with a small scaling factor followed by fine-tuning the network. Zhao et al. [33] propose a compiler aware neural pruning search on YOLOv4 which uses an automatic neural pruning search algorithm that uses a controller and evaluator. The controller is used to select the search space, pre-layer pruning schemes, and prune the model whereas the evaluator evaluates the model accuracy after every pruning step.

#### B. Quantization

Quantization is the process of approximating a continuous signal by a set of discrete symbols or integer values. The discrete set is selected as per the type of quantization such as integer, floating-point, and fixed-point quantization. Quantizing deep learning based object detector models involves converting the baseline 32-bit parameters (weights, activations, biases) to fewer (e.g., 16 or 8) bits, to achieve lower memory footprint, without significantly reducing model accuracy. Fan et al. [34] proposed an 8-bit integer quantization of all the bias, batch normalization, and activation parameters on SSDLite-MobileNetV2. LCDet [35] proposed a fully quantized 8-bit model in which parameters of each layer of a YOLOv2 object detector were quantized to 8-bit fixed point values. To achieve this, they first stored the minimum and maximum value at each layer and then used relative valued to linearly distribute the closest integer value to all the reduced bitwidth weights.

#### C. Knowledge Distillation

Knowledge Distillation involves transferring learned knowledge from a larger model to a smaller, more compact model. A teacher model is first trained for object detection, followed by a smaller student model being trained to emulate the prediction behavior of the teacher model. The goal is to make the student model learn important features to arrive at the predictions that are very close to that of the original model. The resulting student model reduces the computational power and memory footprint compared to the original teacher model. Kang et al. [36] proposed an instance-conditional knowledge decoding module to retrieve knowledge from the teacher network (RetinaNet with a ResNet-101 classifier model as backbone) via query-based attention. They also used a subtask that optimized the decoding module and feature maps to update the student network (RetinaNet with a simpler ResNet-50 model as backbone). Chen et al. [37] proposed a three-step knowledge distillation process on R-CNN with a Resnet-50 model as backbone. The first step used a feature pyramid distillation process to extract the output features that can mimic the teacher network features. They then used these features to remove the output proposal to perform Regional Distillation (RD), enabling the student (RCNN with a much simpler ResNet-18 model as backbone) to focus on the positive regions. Lastly, Logit Distillation (LD) on the output was used to mimic the final output of the teacher network.

**Table 3: Different object detector model optimization techniques and their performance**

| Name | Technique used | Model compression achieved | Object Detector | Latency improvement | Hardware used |
|---|---|---|---|---|---|
| Wang et al. [32] | Pruning | 32.40 % | SSD | 33.61 % | GTX 1080Ti |
| Zhao et al. [33] | Pruning | 93.27 % | YOLOv4 | 80.70 % | Qualcomm Adreno 64 |
| Fan et al. [34] | Quantization | 75 % | SSDLite-MobileNetV2 | 85.67 % | Zynq ZC706 |
| LCDet [35] | Quantization | 77.79 % | YOLOv2 | 13.66 % | Snapdragon 835 |
| Kang et al. [36] | Knowledge Distillation | 37.11 % | RetinaNet | 32.98 % | - |
| Chen et al. [37] | Knowledge Distillation | 83.82 % | RCNN | 7.93 % | - |

## IV. OPEN CHALLENGES AND OPPORTUNITIES

While there has been significant work on effective object detection for AVs, there are significant outstanding challenges that remain to be solved. Here we discuss some of the key challenges and opportunities for future research in the field.

*Neural Architecture Search (NAS):* In recent years, NAS based efforts have gained much attention to automatically determine the best backbone architecture for a given object detection task. Recent works such as NAS-FCOS [38], MobileDets [39], and AutoDets [40] have shown promising results on image classification tasks. Using automated NAS methods can help identify better anchors boxes and backbone networks to improve object detector performance. The one drawback of these efforts is that they take significantly longer to discover the final architecture. More research is needed to devise efficient NAS approaches targeting object detectors.

*Real-time processing:* Object detectors deployed in AV's use video inputs from AV cameras, but the object detectors are typically trained to detect objects on image datasets. Detecting an object on every frame in a video can increase latency of the detection task. Correlations between consecutive frames can help identify the frames that can be used for detecting new objects (while discarding others) and reduce the latency of the model. Creating models that can correlate spatial and temporal relationships between consecutive frames is an open problem. Recent work on real-time object detection [41], [42], has begun to address this problem, but much more work is needed.

*Sensor Fusion:* Sensor fusion is one of the most widely used methods for increasing accuracy of 2D and 3D object detection. Many efforts fuse lidar and RGB images to perform object detection for autonomous driving. But there are very few works that consider fusion data from ultrasonic sensors, radar, or V2X communication. The fusion of data from more diverse sensors is vital to increasing the reliability of the perception system in AVs. Fusing additional sensor data can also increase stability and ensure that the perception system does not fail when one of the sensors fails due to environmental conditions. Recent efforts [43], [44] are beginning to design object detectors that work with data from various sensors, which is a step in the right direction for reliable perception in AVs.

*Time series information:* Most conventional object detection models rely on a CNN-based network for object detection that does not consider time series information. Only a few works, such as [45] and [46], consider multi-frame perception that uses data from the previous and current time instances. Correlating time series information about vehicle dynamics can increase the reliability of the model. Some works such as Sauer et al. [47] and Chen et al. [48] have used time-series data such as steering angle, vehicle velocity, etc. with object detector output to create a closed loop autonomous driving system. Research on combining these efforts with time-series object detector outputs can enable us to make direct driving decisions from these multi-modal models for safer and more reliable driving.

*Semi-supervised object detection:* Supervised machine learning methods which are used in all object detectors today require an annotated dataset to train the detector models. The major challenge in supervised object detection is to annotate data for different scenarios such as, but not limited to, weather conditions, terrain, variable traffic, and location, which is a time-consuming task to ensure improved safety and adaptability of these models in real-world AV driving scenarios. Due to the evolving changes in driving environments, the use of semi-supervised learning for object detection can reduce training time of these models. Some recent efforts, e.g., [49], [50], [51] advocate for performing object detection using semi-supervised transformer models. Due to the high accuracy of transformer-based models, they can yield better performance when detecting object for autonomous driving tasks. Even though transformer-based models yield higher accuracy, deploying them on embedded onboard computers is still a challenge due to their large memory footprint, which requires further investigation.

*Open Datasets:* Object detector model performance can vary due to changing lighting, weather, and other environmental conditions. Data from different weather conditions during training can help fit all the environmental needs to address this problem. Adding new data to accommodate these weather conditions changes when training and testing these models can help overcome this issue. The Waymo open dataset [52] has a wide variety of data that focus on different lighting and weather conditions to overcome this issue. More such open datasets are needed to train reliable object detectors for AVs to ensure robust performance in a variety of environmental conditions.

*Resource Constraints:* Most object detectors have high computational and power overheads when deployed on real hardware platforms. To address this challenge, prior efforts have adapted pruning, quantization, and knowledge distillation techniques (see Section III) to reduce model footprint and decrease the model's computational needs. New approaches for hardware-friendly pruning and quantization, such as recent efforts [53], [54], can be very useful. Techniques to reduce matrix multiplication operations, such as [55], [56], [57] can also speed up object detector execution time. Hardware and software co-design, by combining pruning, quantization, knowledge distillation etc. along with hardware optimization such as parallel factors adjustment, resource allocation etc. also represents an approach to improve object detector efficiency. Results from recent work [58], [59], [60] have been promising, but much more research is needed on these topics.

## V. CONCLUSION

In this article, we discussed the landscape of various object detectors being considered and deployed in emerging AVs, the challenges involved in using these object detectors in AVs, and how the object detectors can be optimized for lower computational complexity and faster inference during real-time perception. We also presented a multitude of open challenges and opportunities to advance the state-of-the-art with object detection for AVs. As AVs are clearly the transportation industry's future, research to overcome these challenges will be crucial to creating a safe and reliable transportation model.

## VI. Acknowledgments

This work was supported by National Science Foundation (NSF), through grant CNS-2132385.

## About the Authors

**Abhishek Balasubramaniam** (abhishek.balasubramaniam@colostate.edu) is a graduate research assistant in the ECE Dept at Colorado State University.

**Sudeep Pasricha** (sudeep@colostate.edu) is a Professor of ECE, CS, and Systems Engineering at Colorado State University.

## References

[1] Automated Vehicles for Safety, NHTSA Report, 2021.
[2] V. Kukkala, et al. "Advanced Driver Assistance Systems: A Path Toward Autonomous Vehicles ", IEEE Consumer Electronics, Sept 2018.
[3] J3016B: Taxonomy and definitions for terms related to driving automation systems for on-road motor vehicles - SAE international, https://www.sae.org/standards/content/j3016_201806.
[4] R. D. Staff, "Navlab: The self-driving car of the '80s," Rediscover the '80s, 2016.
[5] E. D. Dickmanns, Dynamic Vision for perception and control of Motion. 2010.
[6] R. Lawler, Riding shotgun in Tesla's fastest car ever, 2014.
[7] "Drive Me, the world's most ambitious and advanced public autonomous driving experiment, starts today," Volvo Cars Global Media Newsroom, 2016.
[8] Safety report and Whitepapers, Waymo, https://waymo.com/safety/.
[9] S. McEachern, et al. "Cruise founder takes company's first driverless ride on SF Streets: Video," GM Authority, 2021.
[10] L. Jiao, et al. "A Survey of Deep Learning-Based Object Detection," in IEEE Access, vol. 7, pp. 128837-128868, 2019.
[11] N. Dalal, et al. "Histograms of oriented gradients for human detection", Proc. IEEE CVPR, pp. 886-893, Jun. 2005.
[12] P. F. Felzenszwalb, et al. "Object Detection with Discriminatively Trained Part-Based Models," in IEEE TPMAI, vol. 32, no. 9, Sept. 2010.
[13] R. Girshick, et al. "Rich Feature Hierarchies for Accurate Object Detection and Semantic Segmentation," in IEEE CVPR, 2014.
[14] R. Girshick, "Fast R-CNN", Proc. IEEE ICCV, 2015.
[15] S. Ren, et al., "Faster R-CNN: Towards Real-Time Object Detection with Region Proposal Networks," in IEEE TPMAI, vol. 39, no. 6, Jun 2017.
[16] J. Redmon et al., "You Only Look Once: Unified, Real-Time Object Detection," IEEE CVPR, 2016.
[17] J. Redmon and A. Farhadi, "YOLO9000: Better, Faster, Stronger," IEEE CVPR, 2017.
[18] J. Redmon, et al. "Yolov3: An incremental improvement", arXiv preprint arXiv:1804.02767, 2018.
[19] W. Liu et al., "SSD: Single Shot MultiBox Detector", European conference on computer vision. Springer, Cham , 2016.
[20] T.-Y. Lin et al., "Focal Loss for Dense Object Detection", In Proceedings of the IEEE international conference on computer vision, pp. 2980-2988. 2017 .
[21] A. Bochkovskiy et al., "YOLOv4: Optimal Speed and Accuracy of Object Detection", arXiv preprint arXiv:2004.10934, 2020.
[22] Ultralytics, "Ultralytics/yolov5: Yolov5 in PyTorch & ONNX & CoreML & TFLite," GitHub. https://github.com/ultralytics/yolov5.
[23] C.-Y. Wang et al., "You Only Learn One Representation: Unified Network for Multiple Tasks", arXiv preprint arXiv:2105.04206, 2021.
[24] Z. Ge, et al. "Yolox: Exceeding Yolo Series in 2021", arXiv preprint arXiv:2107.08430, 2021.
[25] F. Nobis et al., "A Deep Learning-based Radar and Camera Sensor Fusion Architecture for Object Detection", 2019 Sensor Data Fusion: Trends, Solutions, Applications (SDF). IEEE, 2019.
[26] J. Fang et al., "3D Bounding Box Estimation for Autonomous Vehicles by Cascaded Geometric Constraints and Depurated 2D Detections Using 3D Results", arXiv preprint arXiv:1909.01867, 2019.
[27] M.Simon, et al. "ComplexYOLO: Real-time 3d object detection on point clouds", arXiv:1803.06199, 2018.
[28] M. Simon et al. "Complexer-YOLO: Real-time 3D object detection and tracking on semantic point clouds", Proc. IEEE CVPR Workshops, 2019.
[29] L. H. Wen, et al., "Fast and Accurate 3D Object Detection for Lidar-Camera-Based Autonomous Vehicles Using One Shared Voxel-Based Backbone," in IEEE Access, vol. 9, pp. 22080-22089, 2021.
[30] Lu, Yantao, et al. "RAANet: Range-Aware Attention Network for LiDAR-based 3D Object Detection with Auxiliary Density Level Estimation." arXiv preprint arXiv:2111.09515, 2021.
[31] T. Liang et al., "Pruning and Quantization for Deep Neural Network Acceleration: A Survey", Neurocomputing, 461, 370-403, 2021.
[32] Q. Wang, et al. "Small Object Detection Based on Modified FSSD and Model Compression," arXiv preprint arXiv:2108.10503, 2021.
[33] P. Zhao, et al. "Neural Pruning Search for Real-Time Object Detection of Autonomous Vehicles," ACM/IEEE DAC, 2021.
[34] H. Fan, et al. "A Real-Time Object Detection Accelerator with Compressed SSDLite on FPGA," FPT, 2018.
[35] S. Tripathi, et al. "LCDet: Low-Complexity Fully-Convolutional Neural Networks for Object Detection in Embedded Systems," IEEE CVPRW, 2017.
[36] Z. Kang et al, "Instance-Conditional Knowledge Distillation for Object Detection", Advances in Neural Information Processing Systems, 2021.
[37] R. Chen, et al. "Learning Lightweight Pedestrian Detector with Hierarchical Knowledge Distillation," IEEE ICIP, 2019.
[38] N. Wang, et al. NAS-FCOS: "Efficient Search for Object Detection Architectures". Int J Comput Vis 129, 3299–3312, 2021.
[39] Y. Xiong, et al., "MobileDets: Searching for Object Detection Architectures for Mobile Accelerators," IEEE/CVF CVPR, 2021
[40] Z. Li, et al., "AutoDet: Pyramid Network Architecture Search for Object Detection", Int J Comput Vis 129, pp. 1087–1105, 2021.
[41] Z. Haidi et al. "Real-Time Moving Object Detection in High-Resolution Video Sensing." Sensors, 2020.
[42] D. Xuerui et. al. "TIRNet: Object detection in thermal infrared images for autonomous driving", Applied Intelligence, 2021.
[43] R. O. Chavez-Garcia et al., "Multiple Sensor Fusion and Classification for Moving Object Detection and Tracking," in IEEE TITS, 17(2), Feb. 2016.
[44] H. Cho et. al., "A multi-sensor fusion system for moving object detection and tracking in urban driving environments," IEEE ICRA, 2014.
[45] S. Casas, et al. "IntentNet: Learning to predict intention from raw sensor data," in Proc. 2nd Annu. Conf. Robot Learn., 2018.
[46] W. Luo, et al. "Fast and furious: Real time endto-end 3D detection, tracking and motion forecasting with a single convolutional net," in Proc. IEEE/CVF CVPR, Jun. 2018.
[47] A. Sauer et al. "Conditional affordance learning for driving in urban environments," in Proc. 2nd Annu. Conf. Robot Learn., 2018.
[48] C. Chen et al. "DeepDriving: Learning affordance for direct perception in autonomous driving," in Proc. ICCV, Dec. 2015.
[49] X. Enze et al. "DetCo: Unsupervised Contrastive Learning for Object Detection." Proceedings of the IEEE/CVF International Conference on Computer Vision, 2021.
[50] D. Zhigang et al. "UP-DETR: Unsupervised Pre-training for Object Detection with Transformers." IEEE/CVF CVPR, 2021.
[51] B. Amir et al. "DETReg: Unsupervised Pretraining with Region Priors for Object Detection." arXiv preprint arXiv:2106.04550, 2021.
[52] P. Sun et al., "Scalability in Perception for Autonomous Driving: Waymo Open Dataset," IEEE/CVF CVPR, 2020.
[53] L. Jing et al. "AQD: Towards Accurate Quantized Object Detection." IEEE/CVF CVPR 2021.
[54] K. Sungrae, et al. "Zero-Centered Fixed-Point Quantization with Iterative Retraining for Deep Convolutional Neural Network-Based Object Detectors." IEEE Access 9 (2021): 20828-20839.
[55] Pilipović, R et al., An Approximate GEMM Unit for Energy-Efficient Object Detection. Sensors 2021, 21, 4195.
[56] S. Winograd, Arithmetic complexity of computations, vol. 33 Siam, 1980.
[57] S. Kala et al., "UniWiG: Unified Winograd-GEMM Architecture for Accelerating CNN on FPGAs," IEEE VLSID, pp. 209-214, 2019.
[58] X. Zhang et al., "SkyNet: a Hardware-Efficient Method for Object Detection and Tracking on Embedded Systems", Proceedings of Machine Learning and Systems, 2, 216-229 , 2019.
[59] Y. Zhu et al. "Acceleration of pedestrian detection algorithm on novel C2RTL HW/SW co-design platform." IEEE ICGCS, 2010.
[60] Y. Ma et al. "Algorithm-hardware co-design of single shot detector for fast object detection on FPGAs." IEEE/ACM ICCAD, 2018.